\pdfoutput=1
\documentclass[11pt]{article}

\usepackage[final]{acl}

\usepackage{times}
\usepackage{latexsym}

\usepackage{booktabs}
\usepackage{amssymb}
\usepackage{tikz}
\usepackage{multirow}
\usepackage{hyperref}
\usepackage{colortbl}
\usepackage{fancyvrb}
\usetikzlibrary{positioning, patterns, fit, calc}

\usepackage[T1]{fontenc}
\usepackage[utf8]{inputenc}
\usepackage{microtype}
\usepackage{inconsolata}
\usepackage{graphicx}

\title{Too Long, Didn't Model:\\ Decomposing LLM Long-Context Understanding With Novels}

\author{Sil Hamilton$^{1}$, Rebecca M. M. Hicke$^{2}$, Matthew Wilkens$^{1}$, David Mimno$^{1}$ \\
  $^{1}$Department of Information Science \\
  $^{2}$Department of Computer Science \\
  Cornell University \\
  \texttt{\{srh255,rmh327,wilkens,mimno\}@cornell.edu}}

\begin{document}
\maketitle
\begin{abstract}
Although the context length of large language models (LLMs) has increased to millions of tokens, evaluating their effectiveness beyond needle-in-a-haystack approaches has proven difficult.
We argue that novels provide a case study of subtle, complicated structure and long-range semantic dependencies often over 128k tokens in length.
Inspired by work on computational novel analysis, we release the Too Long, Didn't Model (\textsc{TLDM}) benchmark, which tests a model's ability to report plot summary, storyworld configuration, and elapsed narrative time.
We find that none of seven tested frontier LLMs retain stable understanding beyond 64k tokens.
Our results suggest language model developers must look beyond ``lost in the middle'' benchmarks when evaluating model performance in complex long-context scenarios.
To aid in further development we release the \textsc{TLDM} benchmark together with reference code and data.
\end{abstract}

\section{Introduction}
Large language model (LLM) context lengths have expanded to millions of tokens, theoretically enabling the analysis of long and complicated documents.
However, recent research suggests LLMs do not properly integrate information across long contexts \citep{hsiehruler, karpinska-etal-2024-one} and have difficulty keeping track of order within contexts \citep{merrillIllusionStateStateSpace2024, liHowLanguageModels2025}.
This failure mode has been hard to evaluate.

Current long-context benchmarks like Needle In a Haystack \citep{niah} and Passkey Retrieval \citep{mohtashami2023randomaccess} evaluate a model's minimal ability to access ``lost in the middle'' data, but fail to test long-context \emph{understanding}.
Retrieving one relevant document from a sea of irrelevant documents does not replicate how we expect most users use long contexts: to integrate multiple documents to arrive at complex conclusions.

\begin{figure}
    \center
   \includegraphics[width=1\linewidth]{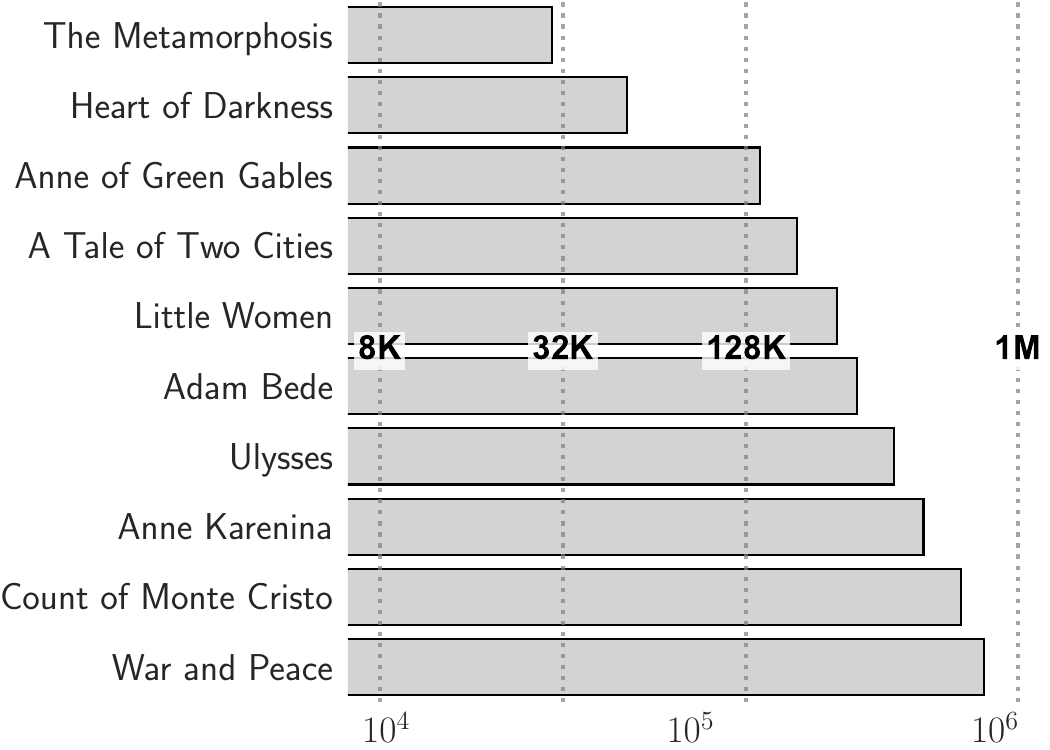}
     \caption{Token lengths of ten popular novels as tokenized by Gemma 3 contrasted with common maximum LLM context lengths (log) as indicated by dotted lines.
     }
      \label{fig:novels}
\end{figure}

We therefore present the Too Long, Didn't Model (\textsc{TLDM}) benchmark: a pipeline for testing long-context LLMs on a set of forty English-language novels ranging in length from <32k to >128k tokens using a suite of narrative understanding tasks including summarization, storyworld description, and narrative time estimation.
Our benchmark circumvents limitations imposed by the need for human annotations by selecting narrative tasks suitable for associative ground construction. 
For example, a novel summary can be decomposed to a series of concatenated chapter summaries.
%This property allows us to reliably generate novel-level ground on a per-model basis, customizing our benchmark to the state-sequencing properties of any particular LLM.
This property allows us to assess  model stability in long-context regimes by measuring the difference between novel-level and chapter-level predictions.\footnote{We make our data and code available \href{https://github.com/srhm-ca/tldm-benchmark}{here.}}

\section{Related Work}
\paragraph{State sequencing.}
It is unknown whether self-supervised machine learning models are able to track state over long inputs.
While theory suggests they can \citep{liuTransformersLearnShortcuts2023, merrillLittleDepthGoes2025}, certain models fail to do so in practice.
\citet{merrillIllusionStateStateSpace2024} show state-space models (e.g. S4 and Mamba) struggle with state-tracking tasks. 
Transformers fare better: researchers have found empirical evidence for state tracking in tasks like entity tracking \cite{li-etal-2021-implicit, kim-schuster-2023-entity}, permutation composition \cite{liHowLanguageModels2025}, Othello \cite{li2023emergent}, and chess \cite{karvonenemergent}.
These results indicate transformers can theoretically process long-context input.

\paragraph{Long-context benchmarking.}
It is popular to  test frontier transformers on long-context inputs by subjecting them to benchmarks such as Needle in a Haystack \cite{niah} and passkey retrieval \cite{mohtashami2023randomaccess}.
These benchmarks test long-context processing by having the model retrieve relevant documents randomly shuffled into sets of irrelevant documents.
Early long-context models (e.g. GPT-4) performed poorly on these tasks, which encouraged language model developers to forefront long-context testing \citep{liuLostMiddleHow2024, liLongcontextLLMsStruggle2024, zhang$infty$BenchExtendingLong2024}.

\paragraph{Literature as benchmark.}
Competent ``lost in the middle'' performance does not mean models can integrate information over long contexts.
Researchers have therefore proposed benchmarks for assessing models on more complicated tasks such as question answering \cite{wang2024leave, yuanLVEvalBalancedLongContext2024}, ``multi-hop reasoning'' \cite{roberts2024needle}, instruction following \cite{baiLongAlignRecipeLong2024}, or all of the above  \cite{chenLongLeaderComprehensiveLeaderboard2025, hsiehruler}. 
Benchmarks have likewise begun to turn to literature as a potential source of natural long-context data \citep{sunLongRangeLanguageModels2021, kimFABLESEvaluatingFaithfulness2024, ahujaFindingFlawedFictions2025}.
One such benchmark, \textsc{NoCha} \cite{karpinska-etal-2024-one} finds open-weight models achieve only  near-random accuracy when querying texts averaging 127k tokens in length --- but \textsc{NoCha} (and other literary benchmarks) fail to provide insight into \textit{when} in the context window models begin to fail to process information.
Our benchmark proposes a unique set of state-oriented narrative understanding tasks to test how models represent state \textit{up to a given point in a novel}.

\section{Methods}
\paragraph{Data.}
We align Project Gutenberg's catalog \cite{hartProjectGutenberg1971} with the MultiHATHI corpus \cite{Hamilton-2023} to identify 6,219 English-only public domain works of fiction averaging 241 pages in length.
We then randomly sample books containing clearly delineated chapters that appear to follow a single narrative.
We keep the first ten suitable texts that fall into each of four length bins: <32k, 32k--64k, 64k--128k, and >128k tokens.\footnote{The length of each text in number of tokens is calculated by applying the Gemma 2 tokenizer to the text file versions of each text with paratextual information  removed.}\footnote{We list all sampled novels in \autoref{sec:appendix-corpus}.}
Our final dataset contains forty novels containing an average of 27 chapters each.

\paragraph{Tasks.}
We deploy three narrative understanding tasks that require processing large amounts of text:

\begin{enumerate}
    \item \textbf{Summarization:} Summarize the narrative with one sentence per chapter.
    \item \textbf{Storyworld description:} Return the last known physical location of every character in the narrative.
    \item \textbf{Narrative time:} Estimate the narrative time passed in hours, days, months, or years.
\end{enumerate}

Each task requires models to extract and report different types of narrative information; the first requires identifying salient plot points, the second requires entity tracking, and the third requires a model of story time independent of narration.\footnote{Task-specific prompts are made available in the appendix.}

\paragraph{Windows of interest.}
Each task-specific prompt instructs the model to perform the task for the first 25\%, 50\%, 75\%, or 100\% of chapters.
This allows us to evaluate whether models can limit results to a subsection of text.
Subsequent results will refer to these subspans as \textit{windows of interest.} 

\paragraph{Text treatments.}
\label{sec:treatments}
\begin{figure}[t]
    \centering
    \begin{tikzpicture}[
        novel_part/.style={
            rectangle, 
            draw=black, 
            fill=gray!20, 
            minimum height=0.4cm, 
            minimum width=0.7cm, 
            text=black, 
            font=\scriptsize, 
            inner sep=1pt
        },
        long_novel/.style={
            rectangle, 
            draw=black, 
            pattern=crosshatch,
            pattern color=black!40,
            minimum height=0.4cm, 
            minimum width=0.7cm,
            text=black
        },
        bracket/.style={
            font=\normalsize, 
            inner sep=0.5pt
        }
    ]
    
    % Fixed starting position
    \def\startX{0}
    \def\labelX{\startX}
    \def\vgap{1.2}
    
    % T1: No novel (exactly as shown)
    \node[font=\footnotesize\bfseries, anchor=west] at (\labelX, 0) {T1: No novel.};
    \node[bracket] (t1_lbracket) at (\startX, -0.5) {[};
    \node[long_novel, anchor=west] (t1_novel) at ([xshift=0.05cm]t1_lbracket.east) {};
    \node[bracket, anchor=west] (t1_rbracket) at ([xshift=0.05cm]t1_novel.east) {]};
    
    % T2: Novel, unaltered
    \node[font=\footnotesize\bfseries, anchor=west] at (\labelX, -\vgap) {T2: Novel, unaltered.};
    \node[bracket] (t2_lbracket) at (\startX, -\vgap-0.5) {[};
    \node[novel_part, anchor=west] (t2_p1) at ([xshift=0.1cm]t2_lbracket.east) {1};
    \node[novel_part, anchor=west] (t2_p2) at ([xshift=0.1cm]t2_p1.east) {2};
    \node[novel_part, anchor=west] (t2_p3) at ([xshift=0.1cm]t2_p2.east) {3};
    \node[novel_part, anchor=west] (t2_p4) at ([xshift=0.1cm]t2_p3.east) {4};
    \node[novel_part, anchor=west] (t2_p5) at ([xshift=0.1cm]t2_p4.east) {5};
    \node[novel_part, anchor=west] (t2_p6) at ([xshift=0.1cm]t2_p5.east) {6};
    \node[bracket, anchor=west] (t2_rbracket) at ([xshift=0.1cm]t2_p6.east) {]};
    
    % T3: Novel, as user message
    \node[font=\footnotesize\bfseries, anchor=west] at (\labelX, -2*\vgap) {T3: Novel, one chapter per user message.};
    \node[bracket] (t3_lbracket_main) at (\startX, -2*\vgap-0.5) {};
    
    % T3 elements with exact spacing
    \node[bracket, anchor=west] (t3_p1_lbracket) at ([xshift=-0.1cm]t3_lbracket_main.east) {[};
    \node[novel_part, anchor=west] (t3_p1) at ([xshift=0.02cm]t3_p1_lbracket.east) {1};
    \node[bracket, anchor=west] (t3_p1_rbracket) at ([xshift=0.02cm]t3_p1.east) {]};
    
    \node[bracket, anchor=west] (t3_p2_lbracket) at ([xshift=0.13cm]t3_p1_rbracket.east) {[};
    \node[novel_part, anchor=west] (t3_p2) at ([xshift=0.02cm]t3_p2_lbracket.east) {2};
    \node[bracket, anchor=west] (t3_p2_rbracket) at ([xshift=0.02cm]t3_p2.east) {]};
    
    \node[bracket, anchor=west] (t3_p3_lbracket) at ([xshift=0.13cm]t3_p2_rbracket.east) {[};
    \node[novel_part, anchor=west] (t3_p3) at ([xshift=0.02cm]t3_p3_lbracket.east) {3};
    \node[bracket, anchor=west] (t3_p3_rbracket) at ([xshift=0.02cm]t3_p3.east) {]};
    
    \node[bracket, anchor=west] (t3_p4_lbracket) at ([xshift=0.13cm]t3_p3_rbracket.east) {[};
    \node[novel_part, anchor=west] (t3_p4) at ([xshift=0.02cm]t3_p4_lbracket.east) {4};
    \node[bracket, anchor=west] (t3_p4_rbracket) at ([xshift=0.02cm]t3_p4.east) {]};
    
    \node[bracket, anchor=west] (t3_p5_lbracket) at ([xshift=0.13cm]t3_p4_rbracket.east) {[};
    \node[novel_part, anchor=west] (t3_p5) at ([xshift=0.02cm]t3_p5_lbracket.east) {5};
    \node[bracket, anchor=west] (t3_p5_rbracket) at ([xshift=0.02cm]t3_p5.east) {]};
    
    \node[bracket, anchor=west] (t3_p6_lbracket) at ([xshift=0.13cm]t3_p5_rbracket.east) {[};
    \node[novel_part, anchor=west] (t3_p6) at ([xshift=0.02cm]t3_p6_lbracket.east) {6};
    \node[bracket, anchor=west] (t3_p6_rbracket) at ([xshift=0.02cm]t3_p6.east) {]};

    % T4: Novel, truncated
    \node[font=\footnotesize\bfseries, anchor=west] at (\labelX, -3*\vgap) {T4: Novel, truncated to window of interest.};
    \node[bracket] (t4_lbracket) at (\startX, -3*\vgap-0.5) {[};
    \node[novel_part, anchor=west] (t4_p1) at ([xshift=0.1cm]t4_lbracket.east) {1};
    \node[novel_part, anchor=west] (t4_p2) at ([xshift=0.1cm]t4_p1.east) {2};
    \node[novel_part, anchor=west] (t4_p3) at ([xshift=0.1cm]t4_p2.east) {3};
    \node[novel_part, anchor=west] (t4_p4) at ([xshift=0.1cm]t4_p3.east) {4};
    \node[bracket, anchor=west] (t4_rbracket) at ([xshift=0.1cm]t4_p4.east) {]};
    
    % T5: Novel, shuffled
    \node[font=\footnotesize\bfseries, anchor=west] at (\labelX, -4*\vgap) {T5: Novel, chapters shuffled.};
    \node[bracket] (t5_lbracket) at (\startX, -4*\vgap-0.5) {[};
    \node[novel_part, anchor=west] (t5_p1) at ([xshift=0.05cm]t5_lbracket.east) {2};
    \node[novel_part, anchor=west] (t5_p2) at ([xshift=0.1cm]t5_p1.east) {1};
    \node[novel_part, anchor=west] (t5_p3) at ([xshift=0.1cm]t5_p2.east) {3};
    \node[novel_part, anchor=west] (t5_p4) at ([xshift=0.1cm]t5_p3.east) {6};
    \node[novel_part, anchor=west] (t5_p5) at ([xshift=0.1cm]t5_p4.east) {5};
    \node[novel_part, anchor=west] (t5_p6) at ([xshift=0.1cm]t5_p5.east) {4};
    \node[bracket, anchor=west] (t5_rbracket) at ([xshift=0.1cm]t5_p6.east) {]};
    
    \end{tikzpicture}
    \caption{The five treatments considered in our study as applied to a six-chapter novel. Brackets indicate user message beginning and end while rectangles indicate chapters. Crosshatches indicate absence of input.}
    \label{fig:treatments}
\end{figure}
We probe for the circumstances in which LLMs fail to process long contexts by permuting all texts with five treatments, presented in \autoref{fig:treatments}.\footnote{We provide example summary responses in \autoref{sec:appendix-gpt}.}
Our first treatment (T1) forgoes the actual text for the novel title and author.\footnote{Note we pass in text author and title for all inputs independent of T1. This condition tests for text memorization.}
Our second treatment (T2) passes in the input text unaltered. 
The third treatment (T3) wraps each chapter in a unique user message, with the intuition being that explicitly delineating chapters could aid models in parsing long inputs. 
Our fourth treatment (T4) truncates the input text to the window of interest.
Our final treatment (T5) randomly shuffles the chapters to test whether models are able to reconstruct narratives from anachronous input text.
We further consider all possible combinations of T3-5 for a total of nine treatments over each input text.

\paragraph{Evaluation.} 
\textsc{TLDM} assesses the stability of model predictions made over long contexts relative to short-context responses.
We do not have human-labeled ground truth, but instead compare individual model performance on short contexts to performance on long contexts.
Contemporary LLMs are often pretrained with a context length of 4096 tokens before being generalized to longer contexts in post-training \citep{abdinPhi4TechnicalReport2024, yangQwen3TechnicalReport2025, suRoFormerEnhancedTransformer2023}.
This token range is approximately the length of the average English novel chapter.\footnote{The mean chapter length in our corpus is 2,845 words or 3,696 Gemma 2 tokens.}
We therefore generate chapter-level outputs independently and concatenate them to create model-specific novel-level predictions.\footnote{We concatenate chapter-level summaries; take the last recorded location of a character across all chapters (recurrently passing in characters from previous chapters to stabilize predictions); and sum per-chapter predictions in seconds.}
We then prompt the model with each text treated as described in \autoref{sec:treatments}.
We finally compute the difference between these full-text predictions and the concatenated short-context output using a similarity heuristic normalized to the range $[0,1]$.\footnote{Semantic similarity for summaries, Jaccard similarity plus semantic similarity for storyworld descriptions, and absolute relative error for time.}

\section{Results}
\begin{table}[t]
    \raggedright
    \setlength{\tabcolsep}{3pt}
    \small
    \begin{tabular}{llrcc}
        \toprule
        \textbf{Dev.} & \textbf{Model} & \textbf{Context} & \textbf{Release} & \textbf{OW} \\
        \midrule
        Meta & Llama 4 Scout & 10,000,000 & 4/2025 & \checkmark \\
        \midrule
        OpenAI & GPT-4.1 & 1,000,000 & 4/2025 & $\times$ \\
        DeepSeek & DeepSeek V3 & 1,000,000 & 2/2025 & \checkmark \\
        Google & Gemini 2.0 Flash & 1,000,000 & 2/2025 & $\times$ \\
        \midrule
        Google & Gemma 3 27b & 128,000 & 4/2025 & \checkmark \\
        Alibaba & Qwen 3 32b & 128,000 & 4/2025 & \checkmark \\
        Mistral & Mistral Small 3.1 & 128,000 & 3/2025 & \checkmark \\
        \bottomrule
    \end{tabular}
    \caption{Comparison of recent large language models sorted by context window size and release date. OW indicates open weights.}
    \label{tab:llm-comparison}
\end{table}
We test seven recent frontier models (\autoref{tab:llm-comparison}) on the \textsc{TLDM} benchmark to evaluate the current state of the art.
All seven models were released in 2025 and support from 128k to 10 million input tokens.
We access GPT-4.1 \citep{openaiIntroducingGPT41API2025} and DeepSeek V3 \citep{deepseekaiDeepSeekV3TechnicalReport2025} via Microsoft Azure, Mistral Small 3.1 \citep{mistralaiteamMistralSmall312025} via the Mistral API, and Gemini 2.0 Flash \citep{teamgeminiGemini20Flash2025} \& Gemma 3 27b \citep{teamgemmaGemma3Technical2025} via the Google AI Studio API. 
We then host Qwen 3 32b \citep{yangQwen3TechnicalReport2025}\footnote{We disable chain-of-thought token sampling for Qwen 3 32b to maintain even footing with the other models.} and Llama 4 Scout \citep{metaLlama4Herd2025} on two Nvidia H200 on AWS.\footnote{We consume a total of \$600 in compute credits across all services.} Values for each length bin are averaged over 10 novels.

\begin{table*}[t]
    \small
    \centering
    \setlength{\tabcolsep}{3.1pt}
    \renewcommand{\arraystretch}{1.1}
    \begin{tabular}{lcccc|c|cccc|c|cccc|c}
        & \multicolumn{5}{c}{\textbf{Summary}} & \multicolumn{5}{c}{\textbf{Storyworld}} & \multicolumn{5}{c}{\textbf{Time}} \\
        \cmidrule(lr){2-6} \cmidrule(lr){7-11} \cmidrule(lr){12-16}
        \textbf{Model} & <32 & 32-64 & 64-128 & >128 & B & <32 & 32-64 & 64-128 & >128 & B & <32 & 32-64 & 64-128 & >128 & B \\
        \midrule
        GPT-4.1 & \cellcolor[rgb]{0.70,0.95,0.70}0.80 & \cellcolor[rgb]{0.68,0.96,0.68}0.81 & \cellcolor[rgb]{0.68,0.96,0.68}0.81 & \cellcolor[rgb]{0.66,0.97,0.66}0.82 & \cellcolor[rgb]{0.93,0.86,0.86}0.27 & \cellcolor[rgb]{0.97,0.83,0.83}0.17 & \cellcolor[rgb]{0.93,0.87,0.87}0.27 & \cellcolor[rgb]{0.97,0.83,0.83}0.16 & \cellcolor[rgb]{0.99,0.81,0.81}0.09 & \cellcolor[rgb]{1.00,0.80,0.80}0.00 & \cellcolor[rgb]{0.82,0.92,0.82}0.58 & \cellcolor[rgb]{0.84,0.91,0.84}0.54 & \cellcolor[rgb]{0.84,0.91,0.84}0.54 & \cellcolor[rgb]{0.91,0.88,0.88}0.38 & \cellcolor[rgb]{0.92,0.87,0.87}0.35 \\
        
        Llama 4 Scout & \cellcolor[rgb]{0.75,0.94,0.75}0.76 & \cellcolor[rgb]{0.74,0.94,0.74}0.77 & \cellcolor[rgb]{0.76,0.93,0.76}0.74 & \cellcolor[rgb]{0.74,0.94,0.74}0.77 & \cellcolor[rgb]{0.93,0.87,0.87}0.29 & \cellcolor[rgb]{0.98,0.81,0.81}0.10 & \cellcolor[rgb]{0.98,0.81,0.81}0.10 & \cellcolor[rgb]{0.99,0.80,0.80}0.07 & \cellcolor[rgb]{1.00,0.80,0.80}0.02 & \cellcolor[rgb]{1.00,0.80,0.80}0.00 & \cellcolor[rgb]{0.83,0.91,0.83}0.55 & \cellcolor[rgb]{0.80,0.93,0.80}0.60 & \cellcolor[rgb]{0.82,0.92,0.82}0.57 & \cellcolor[rgb]{0.81,0.92,0.81}0.58 & \cellcolor[rgb]{0.94,0.85,0.85}0.24 \\
        
        Gemini 2.0 Flash & \cellcolor[rgb]{0.77,0.93,0.77}0.73 & \cellcolor[rgb]{0.76,0.94,0.76}0.75 & \cellcolor[rgb]{0.79,0.92,0.79}0.69 & \cellcolor[rgb]{0.77,0.93,0.77}0.72 & \cellcolor[rgb]{0.93,0.86,0.86}0.27 & \cellcolor[rgb]{0.98,0.82,0.82}0.12 & \cellcolor[rgb]{0.97,0.83,0.83}0.14 & \cellcolor[rgb]{0.99,0.81,0.81}0.09 & \cellcolor[rgb]{0.99,0.80,0.80}0.05 & \cellcolor[rgb]{1.00,0.80,0.80}0.00 & \cellcolor[rgb]{0.83,0.91,0.83}0.55 & \cellcolor[rgb]{0.77,0.93,0.77}0.67 & \cellcolor[rgb]{0.84,0.91,0.84}0.54 & \cellcolor[rgb]{0.79,0.93,0.79}0.63 & \cellcolor[rgb]{0.93,0.86,0.86}0.29 \\
        
        DeepSeek V3 & \cellcolor[rgb]{0.76,0.94,0.76}0.75 & \cellcolor[rgb]{0.70,0.95,0.70}0.80 & \cellcolor[rgb]{0.79,0.92,0.79}0.69 & \cellcolor[rgb]{0.93,0.86,0.86}0.30 & \cellcolor[rgb]{0.94,0.85,0.85}0.24 & \cellcolor[rgb]{0.97,0.83,0.83}0.14 & \cellcolor[rgb]{0.98,0.82,0.82}0.12 & \cellcolor[rgb]{0.99,0.81,0.81}0.08 & \cellcolor[rgb]{1.00,0.80,0.80}0.00 & \cellcolor[rgb]{1.00,0.80,0.80}0.00 & \cellcolor[rgb]{0.86,0.90,0.86}0.50 & \cellcolor[rgb]{0.81,0.92,0.81}0.58 & \cellcolor[rgb]{0.79,0.92,0.79}0.61 & \cellcolor[rgb]{0.88,0.89,0.88}0.45 & \cellcolor[rgb]{0.93,0.86,0.86}0.28 \\
        
        Mistral Small 3.1 & \cellcolor[rgb]{0.70,0.95,0.70}0.80 & \cellcolor[rgb]{0.77,0.93,0.77}0.73 & \cellcolor[rgb]{0.80,0.92,0.80}0.62 & \cellcolor[rgb]{0.95,0.84,0.84}- & \cellcolor[rgb]{0.94,0.85,0.85}0.24 & \cellcolor[rgb]{0.94,0.85,0.85}0.24 & \cellcolor[rgb]{0.96,0.84,0.84}0.19 & \cellcolor[rgb]{0.98,0.82,0.82}0.11 & \cellcolor[rgb]{0.95,0.84,0.84}- & \cellcolor[rgb]{1.00,0.80,0.80}0.01 & \cellcolor[rgb]{0.81,0.92,0.81}0.58 & \cellcolor[rgb]{0.87,0.90,0.87}0.47 & \cellcolor[rgb]{0.81,0.92,0.81}0.59 & \cellcolor[rgb]{0.95,0.84,0.84}- & \cellcolor[rgb]{0.93,0.86,0.86}0.26 \\
        
        Gemma 3 27b & \cellcolor[rgb]{0.77,0.93,0.77}0.72 & \cellcolor[rgb]{0.79,0.92,0.79}0.66 & \cellcolor[rgb]{0.97,0.83,0.83}0.16 & \cellcolor[rgb]{0.95,0.84,0.84}- & \cellcolor[rgb]{0.94,0.85,0.85}0.25 & \cellcolor[rgb]{0.96,0.83,0.83}0.18 & \cellcolor[rgb]{0.93,0.87,0.87}0.25 & \cellcolor[rgb]{1.00,0.80,0.80}0.01 & \cellcolor[rgb]{0.95,0.84,0.84}- & \cellcolor[rgb]{1.00,0.80,0.80}0.00 & \cellcolor[rgb]{0.79,0.93,0.79}0.62 & \cellcolor[rgb]{0.84,0.91,0.84}0.54 & \cellcolor[rgb]{0.86,0.90,0.86}0.50 & \cellcolor[rgb]{0.95,0.84,0.84}- & \cellcolor[rgb]{0.94,0.85,0.85}0.22 \\
        
        Qwen 3 32b & \cellcolor[rgb]{0.73,0.94,0.73}0.78 & \cellcolor[rgb]{0.77,0.93,0.77}0.73 & \cellcolor[rgb]{0.80,0.92,0.80}0.62 & \cellcolor[rgb]{0.95,0.84,0.84}- & \cellcolor[rgb]{0.93,0.86,0.86}0.26 & \cellcolor[rgb]{0.93,0.86,0.86}0.25 & \cellcolor[rgb]{0.99,0.81,0.81}0.06 & \cellcolor[rgb]{1.00,0.80,0.80}0.04 & \cellcolor[rgb]{0.95,0.84,0.84}- & \cellcolor[rgb]{1.00,0.80,0.80}0.00 & \cellcolor[rgb]{0.82,0.92,0.82}0.56 & \cellcolor[rgb]{0.78,0.93,0.78}0.64 & \cellcolor[rgb]{0.82,0.92,0.82}0.56 & \cellcolor[rgb]{0.95,0.84,0.84}- & \cellcolor[rgb]{0.92,0.87,0.87}0.32 \\
        \bottomrule
    \end{tabular}
    \caption{LLM performance comparing full novels (T2) to per-chapter results across different volume lengths, with similarity scores for summaries, storyworlds, and times. Values are normalized to the closed interval $[0,1]$. All reported lengths are in thousands of Gemma 2 tokens. Performance is consistent across models below the 64k--128k length bracket. Compare with random baselines (B) averaged over 1,440 random pairs on a per-model basis.}
    \label{tab:model-performance}
\end{table*}

\paragraph{Full-novel performance.} 
We first report results where the model is asked to analyze the entire input text (whole unaltered novels, treatment 2 in \autoref{tab:model-performance}).
We find that all models exhibit comparable performance when processing volumes with <64k tokens but that performance begins to degrade as book lengths exceed 64k tokens.
Performance degrades at different rates, with summary and storyworld scores dropping faster than time estimate scores. Open-weight models (particularly Gemma 3 27b and Qwen 3 32b) exhibit the steepest decline in performance.
Of the models equipped to process over 128k tokens, we find GPT-4.1 is most consistent across all context lengths.
Llama 4 Scout and Gemini 2.0 Flash are the next most resilient, achieving reliable performance in summary and time estimation over all lengths.
However, no model  performed well in estimating storyworlds, suggesting models grow increasingly inconsistent in their descriptions when processing individual chapters versus whole chapters.
Finally, we find that performance scales with parameter count.

\paragraph{Treatment impact.}
Examining the effect of each treatment on average model performance reveals several trends.
First, increasing novel length decreases model performance across all treatments (excluding title/author only, T1).
Even when we only ask models to analyze a subset of the provided text (the ``window of interest''), the same pattern holds.
In fact, differences in model performance between short and long texts increase as the window of interest increases. 
That is, there is a greater difference in average model performance for <32k and >128k novels when we request analysis of all chapters than when we ask for 25\% of chapters. 
This pattern makes intuitive sense, as the models are forced to consider more text comparatively when examining longer windows of longer texts. 

We see limited evidence of memorization from pre-training.
Passing only a volume's title and author to the models (T1) decreases average model performance by roughly a third on the summary task, and to near zero for storyworld; narrative time estimates are equivalent. 
Similarly, shuffling the chapters of a text (T5) reduces the performance on the summaries and storyworlds for shorter text windows but does not consistently affect time estimates. 
Truncating the texts to the window of interest (T4) and all treatment combinations with T4 improve performance for truncated windows of interest; this effect is strongest for the shortest windows and longest texts. 
In contrast, passing individual chapters as user messages (T3) has little consistent impact on performance. 
Finally, we note that the average model performance when reporting storyworlds falls as the window of interest increases from 25\% of the novel to  100\%.

\section{Discussion and Conclusion}
We present the Too Long, Didn't Model (\textsc{TLDM}) benchmark for long-context understanding. 
We release initial benchmark scores for seven frontier LLMs released in early 2025.
The benchmark includes three narrative understanding tasks: summarization, storyworld reporting, and narrative time estimation.
All require models to infer information over the full text of a novel.
It evaluates models on novels of varying lengths (<32k to >128k), assesses the models' ability to focus on a particular subset of texts, and determines the impact of various experimental treatments (e.g.,\ shuffling chapters).
Assessing models via this benchmark therefore provides a comprehensive understanding of their ability to perform complex long-context analysis.
    
\paragraph{No tested model is perfect at long contexts.}
Our preliminary model evaluations show that, despite having context windows of up to 10M tokens, models' performance declines considerably with longer texts (those above 64k tokens), especially for non-summary tasks.
The true context of these models is still limited for complex understanding tasks.

\paragraph{Model scale benefits long-context understanding.}
We find long-context abilities improve linearly with model size.
This indicates smaller, open-weight models that can be run on laptops continue to perform worse than do larger commercial models.

\paragraph{Text linearity aids long-context models.}
We find long-context models are impacted by document order --- especially so when focussing on limited narrative windows. 
This suggests some models develop particularly inelastic mechanisms for tracking linear narratives. 

\paragraph{Next steps.}
We encourage future researchers to investigate whether mechanistic interpretability could yield answers to questions raised herein.
It would be valuable to determine how models are currently performing long-context narrative analysis and tracking narrative state, and whether their representational mechanisms are similar to those of humans.
Doing so would help researchers develop strategies for better predicting which long-context tasks LLMs are most appropriate for.

\section*{Limitations}
\label{sec:limitations}

There are several key limitations to this work. The first is the lack of true ground truth values. The expense and time needed to produce validated human ground truth for full novel-level annotations means that the \textsc{TLDM} benchmark compares novels only to their own short context performance; thus, while we are able to evaluate how model performance extends comparatively to long contexts, we lack maximally robust assessments of model vs.\ human capabilities. Second, all texts included in the benchmark are in English, meaning we do not evaluate models' multilingual performance. Finally, compute restrictions limit the number and variety of models we are able to evaluate in this paper. 
    
\section*{Acknowledgments}
This work was supported by NEH grant HAA-290374-23, AI for Humanists, granted to Matthew Wilkens and David Mimno.

\bibliography{custom}

\appendix

\section{Corpus Contents}
\label{sec:appendix-corpus}
The books included in the TLDM benchmark are: \\
\: \newline
\textbf{<32k}\\
\textbullet \: \textit{Beasley's Christmas Party} by Booth Tarkington \\
\textbullet \: \textit{The Battle Of The Strong (A Romance of Two \\
\hspace*{2em}Kingdoms): Volume 2} by Gilbert Parker\\
\textbullet \: \textit{The Caxtons: Part 12} by Edward Bulwer-Lytton
\textbullet \: \textit{Godolphin: Volume 5} by Edward Bulwer-Lytton \\
\textbullet \: \textit{The Romance of a Christmas Card} by Kate \\
\hspace*{2em}Douglas Wiggin \\
\textbullet \: \textit{The Story of a China Cat} by Laura Lee Hope \\
\textbullet \: \textit{Better Dead} by J.\ M.\ Barrie \\
\textbullet \: \textit{The Tale of Jasper Jay Tuck-Me-In Tales} by \\
\hspace*{2em}Arthur Scott Bailey \\
\textbullet \: \textit{Our Little Hawaiian Cousin} by Mary Hazelton \\
\hspace*{2em}Wade \\
\textbullet \: \textit{Christmas at Thompson Hall} by Anthony\\
\hspace*{2em}Trollope \\
\newline
\textbf{32k--64k} \\
\textbullet \: \textit{Alexander’s Bridge and The Barrel Organ} by \\
\hspace*{2em}Willa Cather and Alfred Noyes\\
\textbullet \: \textit{Tom Swift and His Undersea Search or The \\
\hspace*{2em}Treasure on the Floor of the Atlantic} by\\
\hspace*{2em}Victor Appleton\\
\textbullet \: \textit{Kilmeny of the Orchard} by Lucy Maud \\
\hspace*{2em}Montgomery\\
\textbullet \: \textit{What Will He Do With It: Book 10} by Edward \\
\hspace*{2em}Bulwer-Lytton\\
\textbullet \: \textit{The Tragedy of the Korosko} by Arthur Conan \\
\hspace*{2em}Doyle\\
\textbullet \: \textit{Dorothy Dainty's Gay Times} by Amy Brooks\\
\textbullet \: \textit{Thistle and Rose: A Story for Girls} by Amy\\
\hspace*{2em}Walton\\
\textbullet \: \textit{Ruth Fielding Homeward Bound: A Red Cross\\
\hspace*{2em}Worker's Ocean Perils} by Alice B. Emerson\\
\textbullet \: \textit{Isla Heron} by Laura E. Richards\\
\textbullet \: \textit{Frank Reade, Jr., Fighting the Terror of the \\
\hspace*{2em}Coast} by Anonymous\\
\newline
\textbf{64k--128k} \\
\textbullet \: \textit{Lost in the Fog} by James De Mille\\
\textbullet \: \textit{Dora Deane; Or, The East India Uncle} by Mary \\
\hspace*{2em}Jane Holmes\\
\textbullet \: \textit{Going Some} by Rex Beach\\
\textbullet \: \textit{A Pirate of Parts} by Richard Neville\\
\textbullet \: \textit{The Backwoodsmen} by Charles G.\ D.\ Roberts\\
\textbullet \: \textit{The Watchers: A Novel} by A.\ E.\ W.\ Mason\\
\textbullet \: \textit{Discourses of Keidansky} by Bernard G.\ Richards\\
\textbullet \: \textit{In Queer Street} by Fergus Hume\\
\textbullet \: \textit{Dick Merriwell's Assurance; Or, In His Brother's \\
\hspace*{2em}Footsteps} by Burt L.\ Standish\\
\textbullet \: \textit{The Earl's Promise, A Novel: Volume 2} by Mrs.\ \\ 
\hspace*{2em}J.\ H.\ Riddell\\
\newline
\textbf{128k+} \\
\textbullet \: \textit{The Moon Pool} by Abraham Merritt\\
\textbullet \: \textit{Under Two Flags} by Ouida\\
\textbullet \: \textit{Born in Exile} by George Gissing\\
\textbullet \: \textit{Esther Waters} by George Moore \\
\textbullet \: \textit{Desert Conquest; or, Precious Waters} by A. M.\\
\hspace*{2em}Chisholm\\
\textbullet \: \textit{The Dust Flower} by Basil King\\
\textbullet \: \textit{Wager of Battle: A Tale of Saxon Slavery in\\
\hspace*{2em}Sherwood Forest} by Henry William Herbert\\
\textbullet \: \textit{Betty Alden: The first-born daughter of the
\hspace*{2em}Pilgrims} by Jane G. Austin\\
\textbullet \: \textit{Perch of the Devil} by Gertrude Atherton \\
\textbullet \: \textit{The Brooklyn Murders} by G. D. H. Cole\\

\section{Example Prompts (T1)}
\label{sec:appendix_prompts}

What follows are three example prompts for the novel-level prediction of the 1898 novel "The Battle of the Strong: A Romance of Two Kingdoms" by Gilbert Parker. Each prompt is run with a new user session.

\paragraph{Summary.}
\begin{verbatim}
Source: "The Battle of the Strong: A Romance
of Two Kingdoms" by Gilbert Parker.

Situation: You were given a narrative. You
will now be given a task about the 
narrative. Complete the task. Keep your 
response brief and to the point.

Task: Summarize the narrative with one 
sentence per chapter. Describe what happens.
Do not reference the narrative itself.

Limit your response to the narrative from
chapter 1 up until, and including, chapter
12.
\end{verbatim}

\paragraph{Storyworld description.}
\begin{verbatim}
Source: "The Battle of the Strong: A Romance
of Two Kingdoms" by Gilbert Parker.

Situation: You were given a narrative. You
will now be given a task about the 
narrative. Complete the task. Keep your 
response brief and to the point.

Task: List each character in the narrative 
and their physical location in the story.

Here are a list of possible characters in 
the narrative: {characters}. The list
might be blank.

If the character in the above list is 
present in the narrative, note their last
location in the narrative. If a character
is in the narrative but is not in the above
list, still note their last location in the
narrative. Be consistent with the name.
Only list characters present in the 
narrative. Only list individuals. Do not
list groups of characters.

Limit your response to the narrative from
chapter 1 up until, and including, chapter
12.
\end{verbatim}

JSON Schema follows.

\paragraph{Narrative time estimation.}
\begin{verbatim}
Source: "The Battle of the Strong: A Romance
of Two Kingdoms" by Gilbert Parker.

Situation: You were given a narrative. You
will now be given a task about the 
narrative. Complete the task. Keep your 
response brief and to the point.

Task: Predict how much time in minutes,
hours, or days elapsed in this 
narrative. Specify an exact number and 
unit. Only respond with a number and a 
unit, e.g. 30 minutes, 6 hours, or 1 day.

Limit your response to the narrative from
chapter 1 up until, and including, chapter
12.
\end{verbatim}

\section{Example GPT-4.1 responses.}
\label{sec:appendix-gpt}
Here we provide example summaries for \textit{Christmas at Thompson Hall} by Anthony Trollope.
These summaries were generated with GPT-4.1.
We specifically provide the (concatenated) ground, and the summaries produced under treatments 2 (the whole unaltered novel), 3 (each chapter passed in as a separate user message), and 5 (chapters randomly shuffled) --- with a full window of interest.

\paragraph{Ground.}
1. During a harsh winter journey to attend a family Christmas gathering, an Englishwoman in a Paris hotel braves confusing corridors and an awkward encounter with staff to retrieve mustard as a remedy for her ailing husband's throat. 

2. Mrs. Brown, after preparing a mustard plaster for her supposedly ill husband in a Paris hotel, accidentally applies it to the throat of a sleeping stranger in the wrong room, panics at her mistake, considers---but ultimately avoids---confessing to the night porter, and returns distraught and chilled to her own room. 

3. Mrs. Brown, desperate to escape a French hotel after mistakenly putting a mustard plaster on a stranger instead of her ailing husband Mr. Brown, anxiously concocts excuses to persuade her reluctant husband to leave early while fearing discovery by the night porter and the injured stranger. 

4. Mrs. Brown, after mistakenly entering a stranger's hotel room at night and inadvertently placing a mustard plaster on his throat instead of her husband's, faces public embarrassment and suspicion but ultimately explains the misunderstanding, allowing her and Mr. Brown to narrowly avoid greater trouble and continue their journey, despite lingering awkwardness with the aggrieved stranger and staff. 

5. A series of misunderstandings and mistaken identity occur when Mrs. Brown accidentally applies a mustard plaster to a stranger, Mr. Jones, at a hotel, leading to awkwardness and eventual reconciliation when it is revealed that Mr. Jones is engaged to Mrs. Brown's sister and all parties meet at Thompson Hall.
 
\paragraph{Treatment 2 (whole input).}

1. During a bitterly cold Christmas journey to England, Mrs. Brown struggles with her husband's alleged illness and obtains mustard from their Paris hotel to treat his throat. 

2. In the dark, Mrs. Brown mistakenly applies the mustard poultice to a stranger, Mr. Jones, instead of her husband, and flees in panic rather than wake or help him. 

3. Distraught and desperate to leave the hotel to avoid a confrontation, Mrs. Brown invents excuses to hurry her husband's departure despite guilt and fear of discovery. 

4. Mrs. Brown is confronted about the incident when the porter discovers her handkerchief in Mr. Jones's room, leading to an embarrassing explanation and apology to Mr. Jones, which allows them to continue their journey. 

5. Upon arrival at Thompson Hall, Mrs. Brown and her husband awkwardly discover that the stranger from Paris, Mr. Jones, is her sister's fiancé, leading to the revelation of her mistake and eventual forgiveness during the family's Christmas celebration.

\paragraph{Treatment 3 (separate user messages).}

1. Anxious to reach her family for Christmas despite her husband's complaints and illness, Mrs. Brown brings him to a cold Paris hotel, where he demands a mustard plaster for his throat. 

2. In the middle of the night, Mrs. Brown becomes lost in the hotel while seeking mustard, then mistakenly applies the plaster to a stranger's throat instead of her husband's. 

3. Discovering her error and unable to face the consequences, Mrs. Brown desperately tries to avoid detection and persuade her husband that they must leave quickly for England. 

4. The Browns are confronted by hotel staff and the porter with evidence of the mishap, forcing Mrs. Brown to confess, after which her husband explains the accident to the injured stranger, Mr. Jones, allowing them to depart. 

5. Unexpectedly reunited at Thompson Hall, it is revealed that Mr. Jones is the fiancé of Mrs. Brown's sister, Jane, and after an awkward Christmas gathering, Mrs. Brown's mistake becomes a family anecdote, ultimately forgiven by all.

\paragraph{Treatment 5 (shuffled chapters).}
Certainly! Here is a one-sentence summary for each of chapters I to V:

Chapter I: On a cold Christmas in Paris, Mrs. Brown ventures through a labyrinthine hotel late at night to obtain mustard for her ailing husband's sore throat.

Chapter II: After a series of mishaps, Mrs. Brown mistakenly enters the wrong room and, believing the sleeping man is her husband, applies a mustard plaster to a stranger's throat.

Chapter III: Realizing her error and fearing exposure, Mrs. Brown slips away from the stranger, lies to her husband about her absence, and anxiously resolves they must escape early the next morning.

Chapter IV: As the Browns try to leave, Mrs. Brown is confronted by hotel staff who have evidence of her midnight blunder, leading to an awkward explanation and a necessary apology to the afflicted Mr. Jones before their delayed departure.

Chapter V: Traveling together by coincidence, the Browns and Mr. Jones arrive at Thompson Hall, where, amidst family gatherings and mounting suspicion, the truth of Mrs. Brown's comical mistake is ultimately revealed and reluctantly forgiven.

\section{Treatment Results}
\label{sec:appendix_treatment}
\begin{figure*}[!ht]
   \includegraphics[width=\linewidth]{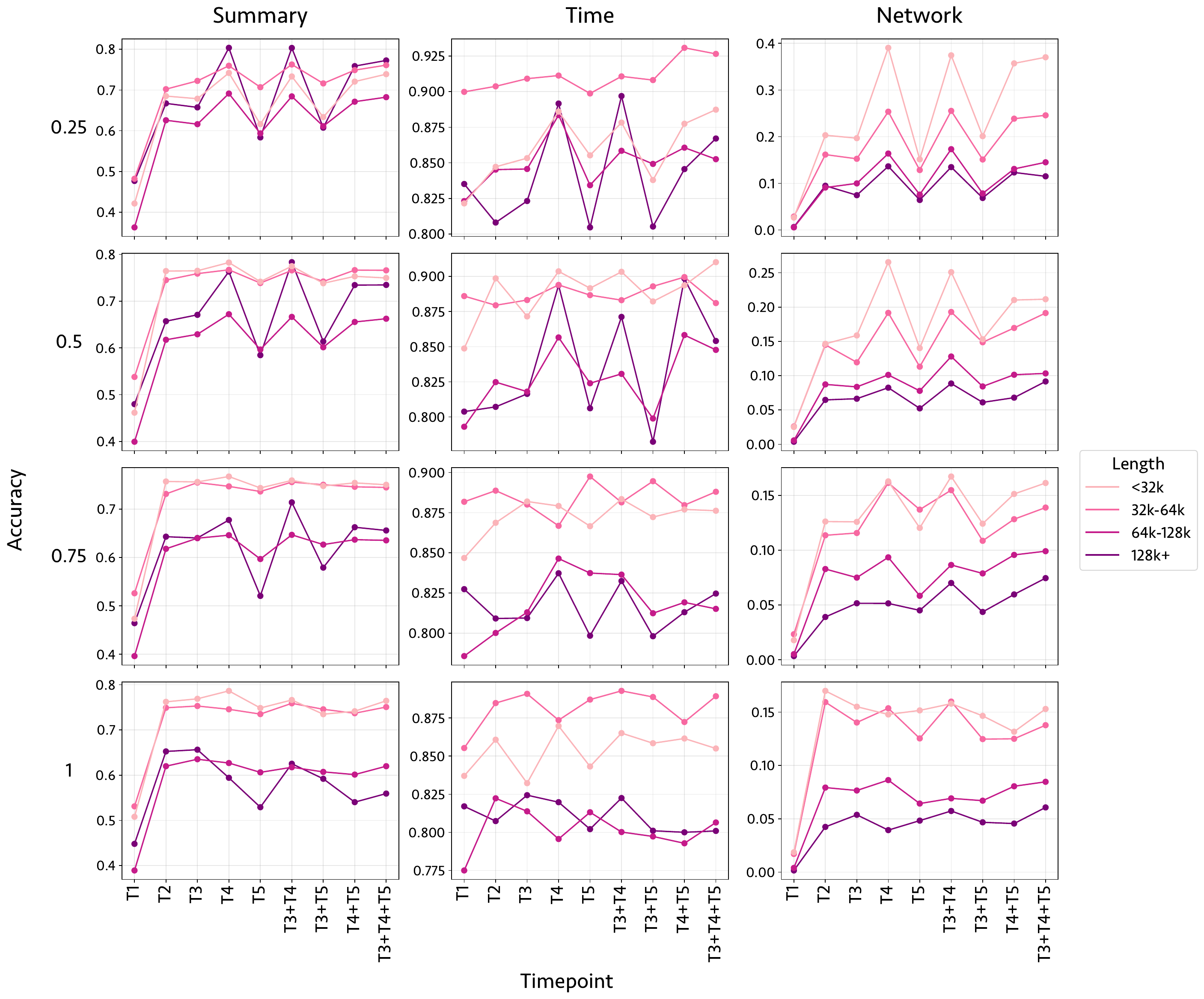}
   \caption{Accuracy scores per task averaged over all models. Values for each window of interest are presented. Note performance consistently degrades as input length increases.}
   \label{fig:experiments}
\end{figure*}

We provide all treated experiment results in \autoref{fig:experiments}.

\end{document}